\documentclass{article}

\usepackage{microtype}
\usepackage{graphicx}
\usepackage{subcaption}
\usepackage{nicefrac}
\usepackage{enumitem}
\usepackage{xspace} 
\usepackage{booktabs} 
\usepackage{amsmath,amssymb,amsthm}

\usepackage{hyperref}

\usepackage[accepted]{mlsys2020}

\mlsystitlerunning{A System for Massively Parallel Hyperparameter Tuning}

\begin{document}

\twocolumn[
\mlsystitle{A System for Massively Parallel Hyperparameter Tuning}

\mlsyssetsymbol{equal}{*}

\begin{mlsysauthorlist}
\mlsysauthor{Liam Li}{cmu}
\mlsysauthor{Kevin Jamieson}{uw}
\mlsysauthor{Afshin Rostamizadeh}{goo}
\mlsysauthor{Ekaterina Gonina}{goo}
\mlsysauthor{Jonathan Ben-Tzur}{det}
\mlsysauthor{Moritz Hardt}{berk}
\mlsysauthor{Benjamin Recht}{berk}
\mlsysauthor{Ameet Talwalkar}{cmu,det}
\end{mlsysauthorlist}

\mlsysaffiliation{cmu}{Carnegie Mellon University}
\mlsysaffiliation{uw}{University of Washington}
\mlsysaffiliation{berk}{University of California, Berkeley}
\mlsysaffiliation{goo}{Google Research}
\mlsysaffiliation{det}{Determined AI}

\mlsyscorrespondingauthor{Liam Li}{me@liamcli.com}

\mlsyskeywords{hyperparameter optimization, automl, distributed machine learning}

\vskip 0.3in

\begin{abstract}
Modern learning models are characterized by large hyperparameter spaces and long training times. These properties, coupled with the rise of parallel computing and the growing demand to productionize machine learning workloads, motivate the need to develop mature hyperparameter optimization functionality in distributed computing settings.  We address this challenge by first introducing a simple and robust hyperparameter optimization algorithm called ASHA, which exploits parallelism and aggressive early-stopping to tackle large-scale hyperparameter optimization problems. Our extensive empirical results show that ASHA outperforms existing state-of-the-art hyperparameter optimization methods; scales linearly with the number of workers in distributed settings; and is suitable for massive parallelism, as demonstrated on a task with 500 workers. 
We then describe several design decisions we encountered, along with our associated solutions, when integrating ASHA in Determined AI's end-to-end production-quality machine learning system that offers hyperparameter tuning as a service.  
\end{abstract}
]

\printAffiliationsAndNotice{}  


\section{Introduction} 
Although machine learning (ML) models have recently achieved dramatic successes in a variety of practical applications, these 
models are highly sensitive to internal parameters, i.e., {\em hyperparameters}.  
In these modern regimes, four trends motivate the need for production-quality systems that  support  massively parallel for hyperparameter tuning:

\textbf{1. High-dimensional search spaces}. 
Models are becoming increasingly complex,
 as evidenced by modern neural networks with dozens of hyperparameters.
 For such complex models with hyperparameters that interact in unknown
ways, a practitioner is forced to evaluate potentially thousands of different hyperparameter settings. 

\textbf{2. Increasing training times}. As datasets grow larger and models become more complex, training a model has become dramatically more expensive, 
 often taking days or weeks on specialized high-performance hardware. 
This trend is particularly onerous in the context of hyperparameter optimization, as a new model must be trained to evaluate each candidate hyperparameter configuration. 

\textbf{3. Rise of parallel computing}. 
The combination of a growing number of hyperparameters and longer training time per model precludes evaluating configurations sequentially; we simply cannot wait years to find a suitable hyperparameter setting.
Leveraging distributed computational resources presents a solution to the increasingly challenging problem of hyperparameter optimization.   

\textbf{4. Productionization of ML}. ML is increasingly driving innovations in industries ranging from autonomous vehicles to scientific discovery to quantitative finance. As ML moves from R\&D to production, ML infrastructure must mature accordingly, with hyperparameter optimization as one of the core supported workloads.

In this work, we address the problem of developing production-quality hyperparameter tuning functionality in a distributed computing setting. Support for massive parallelism is a cornerstone design criteria of such a system and thus a main focus of our work.  

To this end, and motivated by the shortfalls of existing methods, we first introduce \textbf{A}synchronous \textbf{S}uccessive \textbf{H}alving \textbf{A}lgorithm (ASHA), a simple and practical hyperparameter optimization method suitable for massive parallelism that exploits aggressive early stopping.  Our algorithm is inspired by the Successive Halving algorithm (SHA) \citep{karnin2013almost,JamiesonTalwalkar2015}, a theoretically principled early stopping method that allocates more resources to promising configurations.  ASHA is designed for what we refer to as the `large-scale regime,' where to find a good hyperparameter setting, we must evaluate orders of magnitude more hyperparameter configurations than available parallel workers in a small multiple of the wall-clock time needed to train a single model.  

We next perform a thorough comparison of several hyperparameter tuning methods in both the sequential and parallel settings. We focus on `mature' methods, i.e., well-established techniques that have been empirically and/or theoretically studied to an extent that they could be considered for adoption in a production-grade system. In the sequential setting, we compare SHA with Fabolas \citep{fabolas2016}, Population Based Tuning (PBT) \citep{jaderberg2017pbt}, and BOHB \citep{falkner2018bohb}, state-of-the-art methods that exploit partial training.  Our results show that SHA outperforms these methods, which when coupled with SHA's simplicity and theoretical grounding, motivate the use of a SHA-based method in production.  We further verify that SHA and ASHA achieve similar results.  
In the parallel setting, our experiments demonstrate that ASHA addresses the intrinsic issues of parallelizing SHA, scales linearly with the number of workers, and exceeds the performance of PBT, BOHB, and Vizier \citep{vizier2017}, Google's internal hyperparameter optimization  service.

Finally, based on our experience developing ASHA within Determined AI's production-quality 
machine learning system that offers hyperparameter tuning as a service,  
 we describe several systems design decisions and optimizations that we explored as part of the implementation.  We focus on four key considerations: (1) streamlining the user interface to enhance usability; (2) autoscaling parallel training to systematically balance the  tradeoff between lower latency in individual model training and higher throughput in total configuration evaluation; (3) efficiently scheduling ML jobs to optimize multi-tenant cluster utilization; and (4) tracking parallel hyperparameter tuning jobs for reproducibility.


\section{Related Work}
\label{sec:related}

We will first discuss related work that motivated our focus on parallelizing SHA for the large-scale regime.  We then provide an overview of methods for parallel hyperparameter tuning, from which we identify a mature subset to compare to in our empirical studies (Section~\ref{sec:emp}). Finally, we discuss related work on systems for hyperparameter optimization.

\textbf{Sequential Methods}.
Existing hyperparameter tuning methods attempt to speed up the search for a good configuration by either adaptively selecting configurations or adaptively evaluating configurations.  Adaptive configuration selection approaches attempt to identify promising regions of the hyperparameter search space from which to sample new configurations to evaluate \citep{Hutter2011, Snoek2012, Bergstra2011,gpbandit2010}.    However, by relying on previous observations to inform which configuration to evaluate next, these algorithms are inherently sequential and thus not suitable for the large-scale regime, where the number of updates to the posterior is limited.  
In contrast, adaptive configuration evaluation approaches attempt to early-stop
poor configurations and allocate more training ``resources'' to promising
configurations. Previous methods like \citet{GyorgyK11restart,agarwal2012oracle,sabharwal2016} provide theoretical guarantees under strong assumptions on the convergence behavior of intermediate losses.  \cite{krueger2015} relies on a heuristic early-stopping rule based on sequential analysis to terminate poor configurations.  

In contrast, SHA~\citep{JamiesonTalwalkar2015} and Hyperband~\citep{LiJamiesonDeSalvoRostamizadehTalwalkar2015} are adaptive configuration evaluation approaches which do not have the aforementioned drawbacks and have achieved state-of-the-art performance on several empirical tasks.   SHA serves as the inner loop for Hyperband, with Hyperband automating the choice of the early-stopping rate by running different variants of SHA.   While the appropriate choice of early stopping rate
is problem dependent, \citet{LiJamiesonDeSalvoRostamizadehTalwalkar2015}'s empirical results show that aggressive early-stopping works well for a wide variety of tasks.
Hence, we focus on adapting SHA to the parallel setting in Section~\ref{sec:alg}, though we also evaluate the corresponding asynchronous Hyperband method.  

Hybrid approaches combine adaptive configuration selection and evaluation~\cite{Swersky2013Multi,swersky2014freeze,earlystopping2015,fabolas2016}.
\citet{LiJamiesonDeSalvoRostamizadehTalwalkar2015} showed that SHA/Hyperband outperforms SMAC with the learning curve based early-stopping method introduced by \citet{earlystopping2015}.
In contrast, \citet{fabolas2016} reported state-of-the-art performance for Fabolas on several tasks in comparison to Hyperband and other leading methods. However, our results in Section~\ref{ssec:sequential} demonstrate that under an appropriate experimental setup, SHA and Hyperband in fact outperform Fabolas.  Moreover, we note that Fabolas, along with most other Bayesian optimization approaches, can be parallelized using a constant liar (CL) type heuristic \citep{ginsbourger2010, batchLP2016}.  However, the parallel version will underperform the sequential version, since the latter uses a more accurate posterior to propose new points. Hence, our comparisons to these methods are restricted to the sequential setting.

Other hybrid approaches combine Hyperband with adaptive sampling.  For example, \citet{learningcurve2017} combined Bayesian neural networks with Hyperband by first training a Bayesian neural network to predict learning curves and then using the model to select promising configurations to use as inputs to Hyperband.  More recently, \citet{falkner2018bohb} introduced BOHB, a hybrid method combining Bayesian optimization with Hyperband.  They also propose a parallelization scheme for SHA that retains synchronized eliminations of underperforming configurations.  We discuss the drawbacks of this parallelization scheme in Section~\ref{sec:alg} and demonstrate that ASHA outperforms this version of parallel SHA as well as BOHB in Section~\ref{ssec:smallscale}.  We note that similar to SHA/Hyperband, ASHA can be combined with adaptive sampling for more robustness to certain challenges of parallel computing that we discuss in Section~\ref{sec:alg}.

\textbf{Parallel Methods}.
Established parallel methods for hyperparameter tuning include 
PBT \citep{jaderberg2017pbt,pbtsystem} and Vizier \citep{vizier2017}.  PBT is a state-of-the-art hybrid evolutionary approach that exploits partial training to iteratively increase the fitness of a population of models.  In contrast to Hyperband, PBT lacks any theoretical guarantees.  Additionally, PBT is primarily designed for neural networks and is not a general approach for hyperparameter tuning.  We further note that PBT is more comparable to SHA than to Hyperband since both PBT and SHA require the user to set the early-stopping rate via internal hyperparameters.  

Vizier is Google's black-box optimization service with support for multiple hyperparameter optimization methods and early-stopping options. For succinctness, we will refer to Vizier's default algorithm as ``Vizier'' although it is simply one of methods available on the Vizier
platform. While Vizier provides early-stopping rules, the strategies only offer approximately $3\times$ speedup in contrast to the order of magnitude speedups observed for SHA.  We compare to PBT and Vizier in Section~\ref{ssec:smallscale} and Section~\ref{ssec:largescale}, respectively.  

\textbf{Hyperparameter Optimization Systems}.
While there is a large body of work on systems for machine learning, we narrow our focus to systems for hyperparameter optimization.
AutoWEKA \citep{autoweka} and AutoSklearn \citep{Feurer2015} are two established single-machine, single-user systems for hyperparameter optimization.  
Existing systems for distributed hyperparameter optimization include Vizier \citep{vizier2017}, RayTune \citep{liaw2018tune}, CHOPT \cite{chopt} and Optuna \cite{optuna}.
These existing systems provide generic support for a wide range of hyperparameter tuning algorithms; both RayTune and Optuna in fact have support for ASHA.
In contrast, our work focuses on a specific algorithm---ASHA---that we argue is particularly well-suited for massively parallel hyperparameter optimization.  We further introduce a variety of systems optimizations designed specifically to improve the performance, usability, and robustness of ASHA in production environments.  We believe that these optimizations would  directly benefit existing systems to effectively support ASHA, and generalizations of these optimizations could also be beneficial in supporting other  hyperparameter tuning algorithms.

Similarly, we note that \citet{chopt} address the problem of resource management for generic  hyperparameter optimization methods in a shared compute environment, while we focus on efficient resource allocation with adaptive scheduling specifically for ASHA in Section~\ref{ssec:resource_all}.  Additionally, in contrast to the user-specified automated scaling capability for parallel training presented in \citet{xiao2018gandiva}, we propose to automate appropriate autoscaling limits by using the performance prediction framework by \citet{paleo2017}.


\section{ASHA Algorithm}
\label{sec:alg}
\label{async_alg}
We start with an overview of SHA~\citep{karnin2013almost,JamiesonTalwalkar2015} and motivate the need to adapt it to the parallel setting.  Then we present ASHA and discuss how it addresses issues with synchronous SHA and improves upon the original algorithm.

\subsection{Successive Halving (SHA)}
\label{ssec:SHA}

The idea behind SHA (Algorithm~\ref{alg:sha}) is simple: allocate a small budget to each configuration, evaluate all configurations and keep the top $1/\eta$, increase the budget per configuration by a factor of $\eta$, and repeat until the maximum per-configuration budget of $R$ is reached (lines 5--11).  The resource allocated by SHA can be iterations of stochastic gradient descent, number of training examples, number of random features, etc. 

\begin{algorithm}[h]
\begin{algorithmic}
\STATE \textbf{input} number of configurations $n$, minimum resource $r$, maximum resource $R$, reduction factor $\eta$, minimum early-stopping rate $s$
\smallskip
\STATE $s_{\max} = \lfloor\log_\eta(R/r)\rfloor$
\STATE $\textbf{assert } n\geq \eta^{s_{\max} - s}$ so that at least one configuration will be allocated $R$.\\
\STATE $T = {\small \texttt{get\_hyperparameter\_configuration}}(n) $
\STATE {\small \texttt{// All configurations trained for a given $i$ constitute a ``rung.''}}
\FOR{\upshape $i \in \{0, \dots, s_{\max} - s \} $ }
\STATE  $n_i = \lfloor n\eta^{-i}\rfloor$ 
\STATE  $r_i = r\eta^{i+s}$ 
\STATE  $L = {\small\texttt{run\_then\_return\_val\_loss}}(\theta,r_i):\theta\in T$
\STATE  $T ={\small \texttt{top\_k}}(T,L,n_i/\eta)$
\ENDFOR
\STATE \textbf{return} best configuration in $T$
\end{algorithmic}
\caption{Successive Halving Algorithm.}
  \label{alg:sha}
\end{algorithm}

\begin{figure*}[t]
\centering

\begin{subfigure}{0.48\textwidth}
\centering
\includegraphics[width=0.6\textwidth,trim=10 0 10 0]{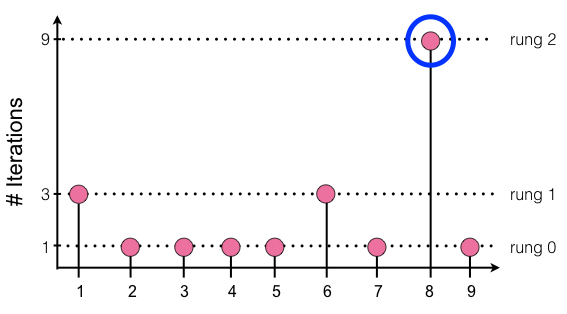}
\subcaption{Visual depiction of the promotion scheme for bracket $s=0$.}
\end{subfigure}
\begin{subfigure}{0.48\textwidth}
\centering

{\begin{tabular}[b]{lcccr}
bracket $s$ & rung $i$ & $n_i$ & $r_i$ & total budget\\ \hline
0 & 0 & 9 & 1 & 9\\
& 1 & 3  & 3 & 9\\
& 2 & 1 & 9 & 9\\ \hline
1 & 0 & 9 & 3 & 27 \\
& 1 & 3 & 9 & 27 \\  \hline
2 & 0 & 9 & 9 & 81 \\

\end{tabular}}
\subcaption{Promotion scheme for different brackets $s$. }
\end{subfigure}
\caption{\textbf{Promotion scheme for  SHA} with $n=9$, $r=1$, $R=9$, and $\eta=3$.
 }
\label{tab:sha}
\end{figure*} 

SHA requires the number of configurations $n$, a minimum resource $r$, a maximum resource $R$, a reduction factor $\eta\geq 2$, and a minimum early-stopping rate $s$.  Additionally, the $\texttt{get\_hyperparameter\_configuration}(n)$ subroutine returns $n$  configurations sampled randomly from a given search space; and the
$\texttt{run\_then\_}$
$\texttt{return\_val\_loss}(\theta, r)$ subroutine
returns the validation loss after training the model with the 
hyperparameter setting $\theta$ and for $r$ resources.
 For a given early-stopping rate $s$, a minimum resource of $r_0 = r\eta^s$ will be allocated to each configuration.
Hence, lower $s$ corresponds to more aggressive early-stopping, with $s=0$ prescribing a minimum resource of $r$.
We will refer to SHA with different values of $s$ as {\em brackets} and, within a bracket,
we will refer to each round of promotion as a {\em rung} with the base rung numbered 0 and increasing.  
Figure~\ref{tab:sha}(a) shows the rungs for bracket 0 for an example setting with $n=9, r=1,R=9,$ and $\eta=3$,  while Figure~\ref{tab:sha}(b) shows how resource allocations change for different brackets $s$. Namely, the starting budget per configuration $r_0 \le R$ increases by a factor of $\eta$ per increment of $s$.  Hence, it takes more resources to explore the same number of configurations for higher $s$.  Note that for a given $s$, the same budget is allocated to each rung but is split between fewer configurations in higher rungs. 

Straightforward ways of parallelizing SHA are not well suited for the parallel regime.
We could consider the embarrassingly parallel approach
of running multiple instances of SHA, one on each worker.
However, this strategy is not well suited for the large-scale regime,
where we would like results in little more than the time to train one configuration. To see this, assume that training time for a configuration scales linearly with the allocated resource and $time(R)$ represents the time required to train a configuration for the maximum resource $R$. In general, for a given bracket $s$, the minimum
time to return a configuration trained to completion is $(\log_\eta(R/r) - s + 1)\times time(R)$, where $\log_\eta(R/r)-s+1$ counts the number of rungs.  For example, consider Bracket 0 in the toy example in Figure~\ref{tab:sha}. The time needed to return a fully trained configuration is $3\times time(R)$, since there are three rungs and each rung is allocated $R$ resource. In contrast, as we will see in the next section, our parallelization scheme for SHA can return an answer in just $time(R)$.  

Another naive way of parallelizing SHA is to distribute the
training of the $n/\eta^k$ surviving configurations on each rung $k$ as is done by \citet{falkner2018bohb} and add brackets when there are no jobs available in existing brackets.  We will refer to this method as ``synchronous'' SHA.
The efficacy of this strategy is severely hampered by two issues: (1) SHA's synchronous nature is sensitive to stragglers and dropped jobs as every configuration within a rung must complete before proceeding to the next rung, and (2) the estimate of the top $1/\eta$ configurations for a given early-stopping rate does not improve as more brackets are run since promotions are performed independently for each bracket.  We demonstrate the susceptibility of synchronous SHA to stragglers and dropped jobs on simulated workloads in Appendix~\ref{appendix:comparison}.

\subsection{Asynchronous SHA (ASHA)}
We now introduce ASHA as an effective technique to parallelize  SHA,
leveraging asynchrony to mitigate stragglers and maximize parallelism.
Intuitively, ASHA promotes configurations to the next
rung whenever possible instead of waiting for a rung 
to complete before proceeding to the next rung.  Additionally, 
if no promotions are possible, ASHA simply
adds a configuration to the base rung,
so that more configurations can be promoted to the upper rungs. 
 ASHA is formally defined in Algorithm~\ref{alg:async}.  Given its asynchronous nature it does not require the
user to pre-specify the number of configurations to evaluate, but it otherwise requires the same inputs as SHA.
Note that the $\texttt{run\_then\_return\_val\_loss}$ subroutine in $\texttt{ASHA}$ is asynchronous and the code execution continues after the job is passed to the worker. ASHA's promotion scheme is laid out in the \texttt{get\_job} subroutine.  

\begin{algorithm}[h]
\begin{algorithmic}
\STATE \textbf{input} {minimum resource $r$, maximum resource $R$, reduction factor $\eta$, minimum early-stopping rate $s$} 
\smallskip
\FUNCTION{{\small \texttt{ASHA}}()}
\REPEAT
\FOR{\upshape for each free worker}
\STATE $(\theta, k) = {\small \texttt{get\_job}}()$ \\
\STATE ${\small \texttt{run\_then\_return\_val\_loss}}(\theta, r\eta^{s+k})$ 
\ENDFOR
\FOR{\upshape completed job ($\theta$, $k$) with loss $l$} 
\STATE Update configuration $\theta$ in rung $k$ with loss $l$.
\ENDFOR
\UNTIL{desired}
\ENDFUNCTION

\smallskip
\FUNCTION{{\small \texttt{get\_job}}()}
\STATE {\small $\texttt{// Check if there is a promotable config.}$}
\FOR{$k=\lfloor\log_\eta(R/r)\rfloor-s-1,\dots,1,0$} 
\STATE candidates $={\small \texttt{top\_k}}(\text{rung } k, |\text{rung } k|/\eta)$ 
\STATE promotable $=\{t \in \text{candidates}:t \text{ not promoted} \}$ 
\IF{\upshape $|\text{promotable}|>0$}
\STATE \textbf{return} \upshape $\text{promotable}[0], k + 1$
\ENDIF
\STATE {\small \texttt{// If not, grow bottom rung.}}
\STATE Draw random configuration $\theta$. 
\STATE \textbf{return} $\theta, 0$
\ENDFOR
\ENDFUNCTION
\end{algorithmic}
\caption{Asynchronous Successive Halving (ASHA)}

  \label{alg:async}
\end{algorithm}

ASHA is well-suited for the large-scale regime, where wall-clock time is constrained to a small
multiple of the time needed to train a single model.  For ease of comparison with SHA, assume training time scales linearly with the resource.  Consider the example of Bracket 0 shown in Figure~\ref{tab:sha}, and assume we can run ASHA with 9 machines. Then ASHA returns a fully trained configuration in $\nicefrac{13}{9}\times time(R)$, since 9 machines are sufficient to promote configurations to the next rung in the same time it takes to train a single configuration in the rung.  Hence, the training time for a configuration in rung 0 is $\nicefrac{1}{9}\times time(R)$, for rung 1 it is $\nicefrac{1}{3}\times time(R)$, and for rung 2 it is $time(R)$. 
In general, $\eta^{\log_\eta(R) - s}$ machines are needed to advance a configuration to the next rung in the same time it takes to train a single configuration in the rung, and it takes 
$\eta^{s + i -\log_\eta(R)}\times time(R)$ to train a configuration in rung $i$. Hence, ASHA can return 
a configuration trained to completion in time
\[\bigg ( \sum_{i=s}^{\log_\eta(R)}\eta^{i-\log_\eta(R)} \bigg )\times time(R) \leq 2\,time(R).\]
Moreover, when training is iterative, ASHA can return an answer in $time(R)$, since
incrementally trained configurations can be checkpointed and resumed.

Finally, since Hyperband simply runs multiple SHA brackets, we can asynchronously parallelize Hyperband by either running multiple brackets of  ASHA or looping through brackets of ASHA sequentially as is done in the original Hyperband.  We employ the latter looping scheme for asynchronous Hyperband in the next section.  

\subsection{Algorithm Discussion}
\label{algo_discussion}

ASHA is able to 
remove the bottleneck associated with synchronous promotions by incurring a small number of incorrect promotions, i.e. configurations that were promoted early on but are not in the top $1/\eta$ of configurations in hindsight.  
By the law of large numbers, we expect to erroneously promote a vanishing fraction of configurations in each rung as the number of configurations grows.
Intuitively, in the first rung with $n$ evaluated configurations, the number 
 of mispromoted configurations is roughly $\sqrt{n}$, since the process
 resembles the convergence of an empirical cumulative
 distribution function (CDF) to its expected value \citep{dkw1956}. 
For later rungs, although the
configurations are no longer  i.i.d.\ since they were advanced 
based on the empirical CDF from the rung below,  we expect
this dependence to be weak.  

We further note that ASHA improves upon SHA in two ways. 
First, \citet{LiJamiesonDeSalvoRostamizadehTalwalkar2015} discusses two SHA variants: finite horizon (bounded resource $R$ per configuration) and infinite horizon (unbounded resources $R$ per configuration).  ASHA consolidates these settings into one algorithm. In Algorithm~\ref{alg:async}, we do not promote configurations that have been trained for $R$, thereby restricting the number of rungs.  However, this algorithm trivially generalizes to the infinite horizon; we can remove this restriction so that the maximum resource per configuration increases naturally as configurations 
are promoted to higher rungs.  In contrast, SHA does not naturally extend to the infinite horizon setting, as it relies on the doubling trick and must rerun brackets with larger budgets to increase the maximum resource. 
 
Additionally, SHA does not return an output until a single bracket completes. In the finite horizon this means that there is a constant interval of $(\text{\# of rungs}\times time(R))$ between receiving outputs from SHA.  In the infinite horizon this interval doubles between outputs.
In contrast, ASHA grows the bracket incrementally instead of in fixed budget intervals.  To further reduce latency, ASHA uses intermediate losses to determine the current best performing configuration, as opposed to only considering the final SHA outputs.


\section{Empirical Evaluation}
\label{sec:emp}
We first present results in the sequential setting to justify our choice of focusing on SHA and to compare SHA to ASHA. 
We next evaluate ASHA in parallel environments 
 on three benchmark tasks.  

\subsection{Sequential Experiments}
\label{ssec:sequential}

We benchmark Hyperband and SHA against PBT, BOHB (synchronous SHA with Bayesian optimization as introduced by \citet{falkner2018bohb}), and Fabolas, and examine the relative performance of SHA versus ASHA and Hyperband versus asynchronous Hyperband.  As mentioned previously, asynchronous Hyperband loops through brackets of ASHA with different early-stopping rates.

\begin{figure}[h]
\centering
\begin{subfigure}{0.48\textwidth}
\centering
\includegraphics[width=0.9\textwidth,trim=10 0 10 0,page=2]{single_machine.pdf}
\end{subfigure}
\begin{subfigure}{0.48\textwidth}
\centering
\includegraphics[width=0.9\textwidth,trim=10 0 10 0,page=1]{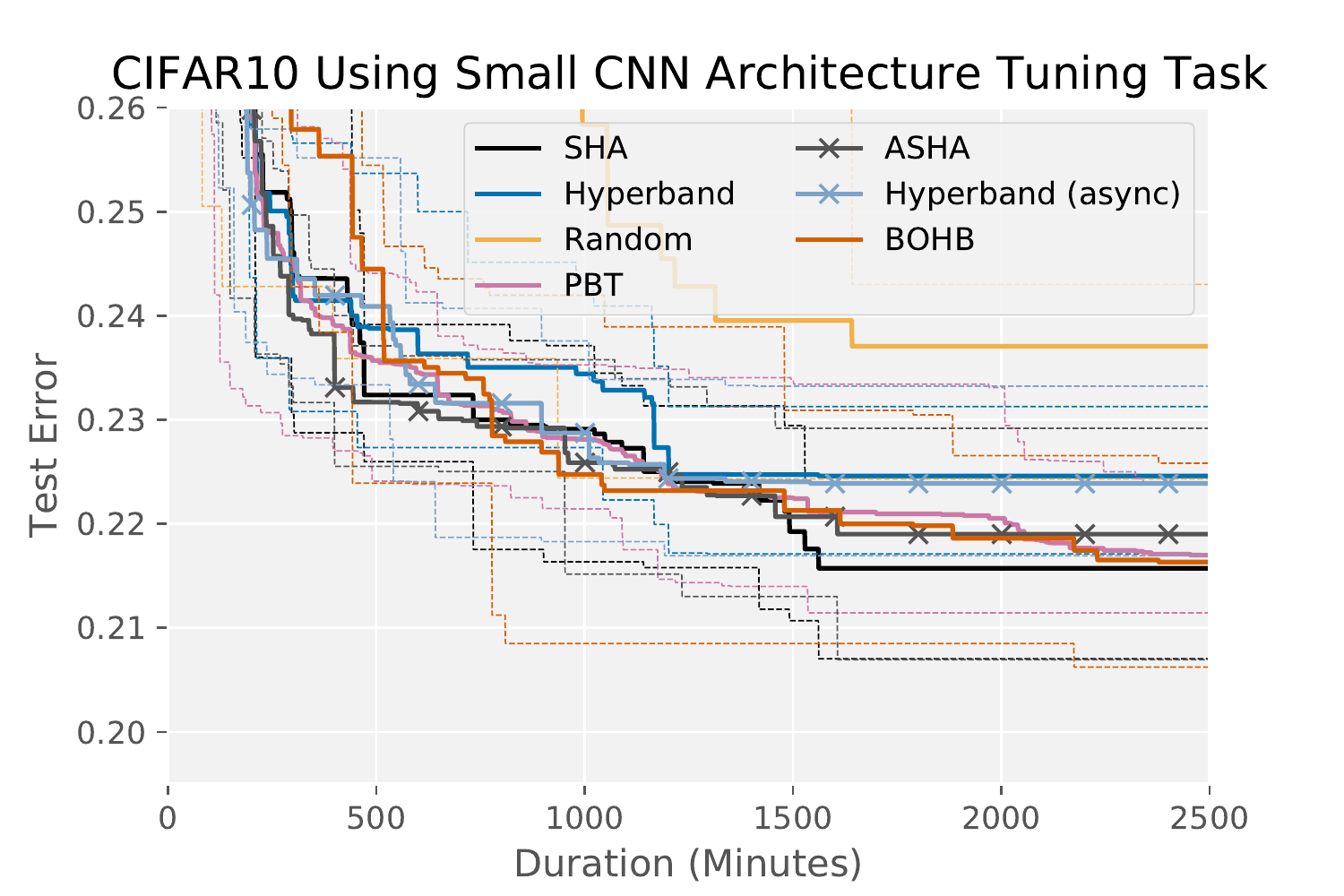}
\end{subfigure}
\caption{\textbf{Sequential experiments} (1 worker).  Average across 10 trials is shown for each hyperparameter optimization method.  Gridded lines represent top and bottom quartiles of trials.} \label{fig:sequential}
\end{figure} 

We compare ASHA against PBT, BOHB, and synchronous SHA on two benchmarks for CIFAR-10: (1) tuning a convolutional neural network (CNN) with the cuda-convnet architecture and the same search space as \cite{hyperband}; and (2) tuning a CNN architecture with varying number of layers, batch size, and number of filters.    The details for the search spaces considered and the settings we used for each search method can be found in Appendix~\ref{appendix:pbt}.  Note that BOHB uses SHA to perform early-stopping and differs only in how configurations are sampled; while SHA uses random sampling, BOHB uses Bayesian optimization to adaptively sample new configurations.  In the following experiments, we run BOHB using the same early-stopping rate as SHA and ASHA instead of looping through brackets with different early-stopping rates as is done by Hyperband.  

The results on these two benchmarks are shown in Figure~\ref{fig:sequential}.  On benchmark 1, Hyperband and all variants of SHA (i.e., SHA, ASHA, and BOHB) outperform PBT by $3\times$.  On benchmark 2, while all methods comfortably beat random search, SHA, ASHA, BOHB and PBT performed similarly and slightly outperform
Hyperband and asynchronous Hyperband. This last observation (i) corroborates
the results in \citet{hyperband}, which found that the brackets with the most aggressive early-stopping rates performed the best; and (ii) follows from the discussion in Section~\ref{sec:related} noting that PBT is more similar in spirit to SHA than Hyperband, as PBT / SHA both require user-specified early-stopping rates (and are more aggressive in their early-stopping behavior in these experiments).
We observe that SHA and ASHA are competitive with BOHB, despite the adaptive sampling scheme used by BOHB.  Additionally, for both tasks, introducing asynchrony does not consequentially impact the performance of ASHA (relative to SHA) or asynchronous Hyperband (relative to Hyperband).  This not surprising; as discussed in Section~\ref{algo_discussion}, we expect the number of ASHA mispromotions to be square root in the number of configurations.  

Finally, due to the nuanced nature of the evaluation framework used by~\citet{fabolas2016}, we present our results on 4 different benchmarks comparing Hyperband to Fabolas in Appendix~\ref{ssec:fabolas}.  In summary, our results show that Hyperband, specifically the first bracket of SHA, tends to outperform Fabolas while also exhibiting lower variance across experimental trials.  

\subsection{Limited-Scale Distributed Experiments }
\label{ssec:smallscale}
We next compare ASHA to synchronous SHA, the parallelization scheme discussed in Section~\ref{ssec:SHA}; BOHB; and PBT on the same two tasks. 
For each experiment,  we run each search method with 25 workers
for 150 minutes. We use the same setups for ASHA and PBT as in the previous section.  We run synchronous SHA and BOHB with default settings and the same $\eta$ and early-stopping rate as ASHA.  

\begin{figure}[t]
\centering
\begin{subfigure}{0.48\textwidth}
\centering
\includegraphics[width=0.9\textwidth,page=1,trim=10 0 10 0]{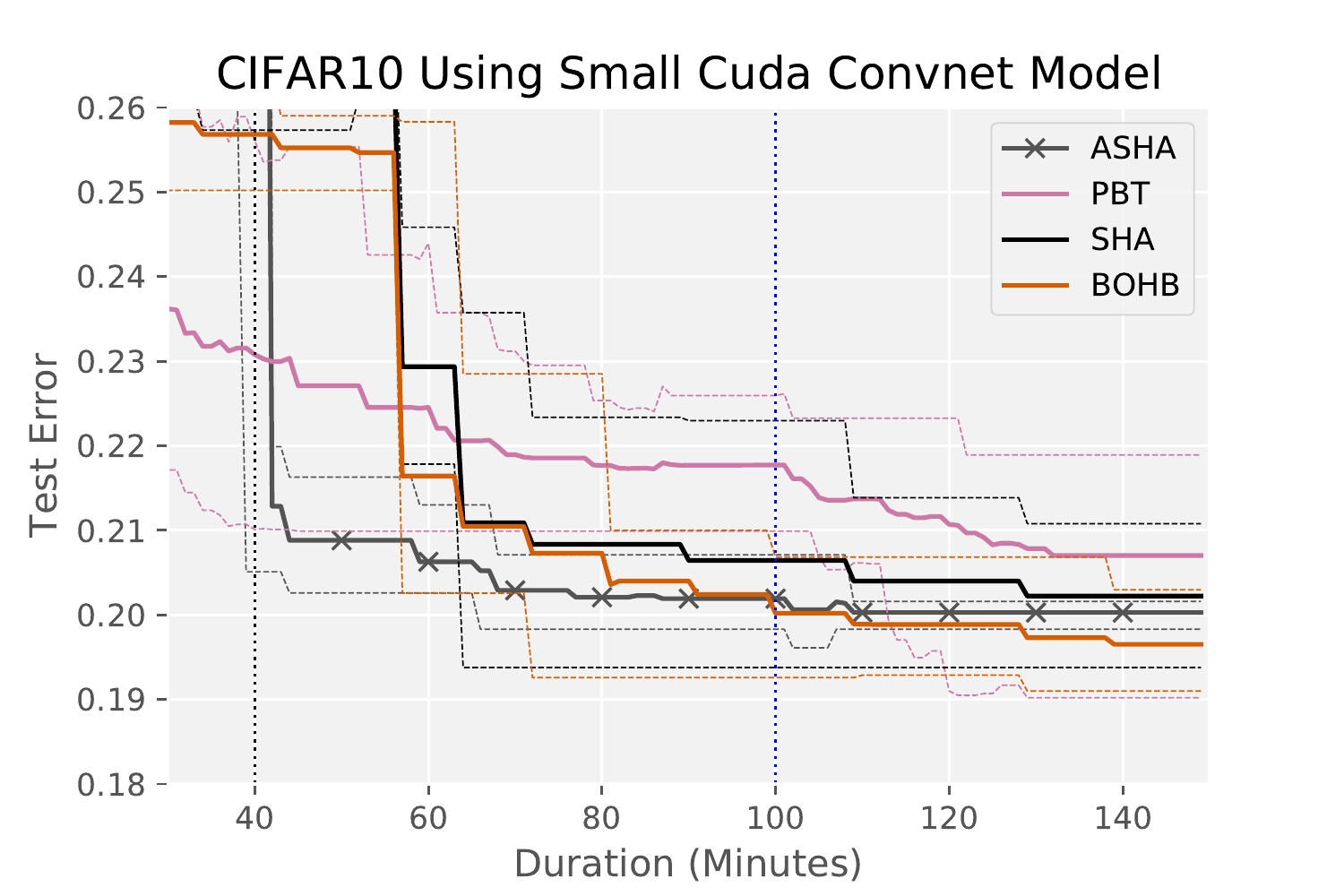}
\end{subfigure}
\begin{subfigure}{0.48\textwidth}
\centering
\includegraphics[width=0.9\textwidth,page=2,trim=10 0 10 0]{parallel_charts.pdf}
\end{subfigure}
\caption{\textbf{Limited-scale distributed experiments} with 25 workers.  For each searcher, the average test error across 5 trials is shown in each plot.  The light dashed lines indicate the min/max ranges.  The dotted black line represents the time needed to train the most expensive model in the search space for the maximum resource $R$.  The dotted blue line represents the point at which 25 workers in parallel have performed as much work as a single machine in the sequential experiments (Figure~\ref{fig:sequential}).}
 \label{fig:smallparallel}
\end{figure}

Figure~\ref{fig:smallparallel} shows the average test error across 5 trials for each search method. 
On benchmark 1, ASHA evaluated over 1000 configurations in just over 40 minutes with 25 workers and found a good configuration (error rate below 0.21) in approximately the time needed to train a single model, whereas it took ASHA nearly 400 minutes to do so in the sequential setting  (Figure~\ref{fig:sequential}). 
Notably, we only achieve a 10$\times$ speedup on 25 workers due to the relative simplicity of this task, i.e.,  
it only required evaluating a few hundred configurations to identify a good one in the sequential setting.  In contrast, for the more difficult search space used in benchmark 2, we observe linear speedups with ASHA, as the $\sim 700$ minutes needed in the sequential setting (Figure~\ref{fig:sequential}) to reach a test error below $0.23$ is reduced to under 25 minutes in the distributed setting.

Compared to synchronous SHA and BOHB, ASHA finds a good configuration $1.5\times$ as fast on benchmark 1 while BOHB finds a slightly better final configuration.  On benchmark 2, ASHA performs significantly better than synchronous SHA and BOHB due to the higher variance in training times between configurations (the average time required to train a configuration on the maximum resource $R$ is 30 minutes with a standard deviation of 27 minutes), which exacerbates the sensitivity of synchronous SHA to stragglers (see Appendix~\ref{appendix:comparison}). BOHB actually underperforms synchronous SHA on benchmark 2 due to its bias towards more computationally expensive configurations, reducing the number of configurations trained to completion within the given time frame.  

We further note that ASHA outperforms PBT on benchmark 1; in fact the minimum and maximum range for ASHA across 5 trials does not overlap with the average for PBT.  On benchmark 2, PBT slightly outperforms asynchronous Hyperband and performs comparably to ASHA.  However, note that the ranges for the searchers share large overlap and the result is likely not significant. Overall, ASHA outperforms PBT, BOHB and SHA on these two tasks.  This improved performance, coupled with the fact that it is a more principled and general approach than either BOHB or PBT (e.g., agnostic to resource type and robust to hyperparameters that change the size of the model), further motivates its use for the large-scale regime.

\subsection{Tuning Neural Network Architectures}
\label{ssec:nas}
Motivated by the emergence of neural architecture search (NAS) as a specialized hyperparameter optimization problem, we evaluate ASHA and competitors on two NAS benchmarks: (1) designing convolutional neural networks (CNN) for CIFAR-10 and (2) designing recurrent neural networks (RNN) for Penn Treebank \citep{ptb}.  We use the same search spaces as that considered by \citet{liu2019darts} (see Appendix~\ref{app:nas} for more details).  

\begin{figure}[h]
\centering
\begin{subfigure}{0.48\textwidth}
\centering
\includegraphics[width=0.9\textwidth,trim=10 0 10 0]{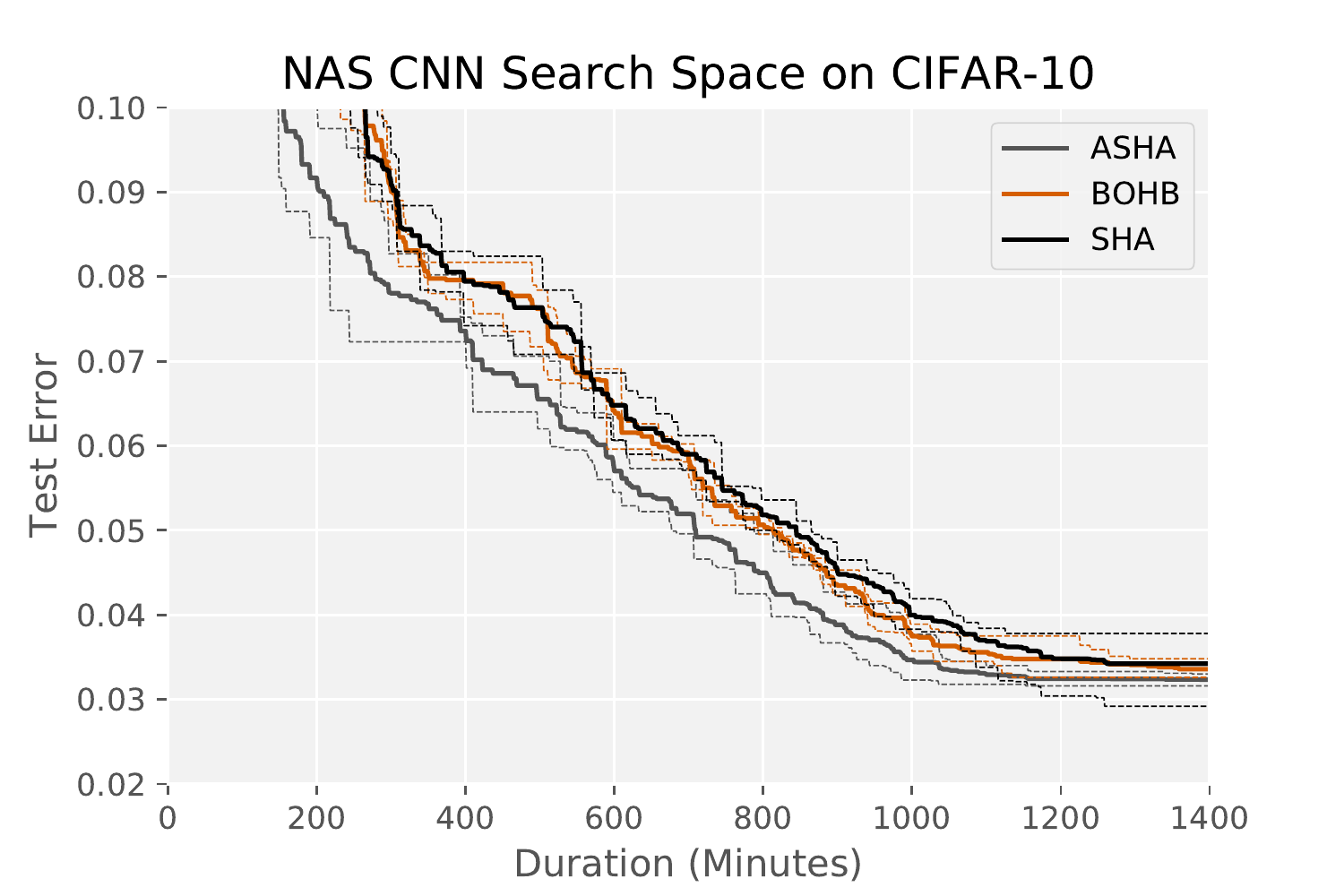}
\end{subfigure}
\begin{subfigure}{0.48\textwidth}
\centering
\includegraphics[width=0.9\textwidth,trim=10 0 10 0]{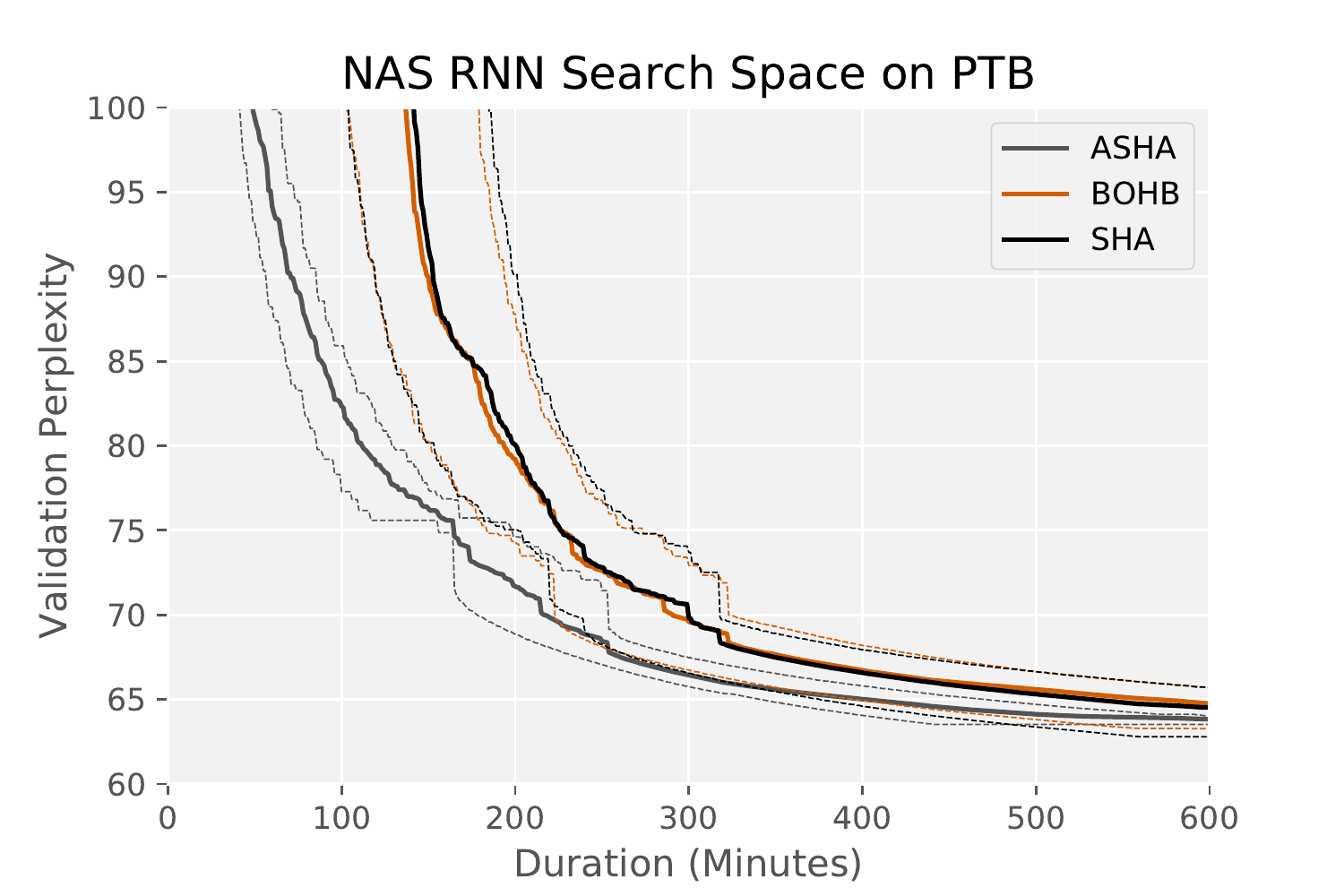}
\end{subfigure}
\caption{\textbf{Tuning neural network architectures} with 16 workers.  For each searcher, the average test error across 4 trials is shown in each plot.  The light dashed lines indicate the min/max ranges.  }
 \label{fig:nas}
\end{figure}

For both benchmarks, we ran ASHA, SHA, and BOHB on 16 workers with $\eta=4$ and a maximum resource of $R=300$ epochs.  The results in Figure~\ref{fig:nas} shows ASHA outperforms SHA and BOHB on both benchmarks.  Our results for CNN search show that ASHA finds an architecture with test error below $10\%$ nearly twice as fast as SHA and BOHB. ASHA also finds final architectures with lower test error on average: $3.24\%$ for ASHA vs $3.42\%$ for SHA and $3.36\%$ for BOHB. Our results for RNN search show that ASHA finds an architecture with validation perplexity below 80 nearly trice as fast as SHA and BOHB and also converges an architecture with lower perplexity: 63.5 for ASHA vs 64.3 for SHA and 64.2 for BOHB. 
Note that vanilla PBT is incompatible with these search spaces since it is not possible to warmstart training with weights from a different architecture.  We show ASHA outperforms PBT in addition to SHA and BOHB on an additional search space for LSTMs in Appendix~\ref{ssec:sota_lstm}.

\subsection{Tuning Large-Scale Language Models}
\label{ssec:largescale}
In this experiment, we increase the number of workers to 500 to evaluate ASHA for massively parallel hyperparameter tuning.  Our search space is constructed based off of the LSTMs considered in \citet{zaremba2014}, with the largest model in our search space matching their large LSTM (see Appendix~\ref{appendix:largescale}).  
For ASHA, 
we set $\eta=4$,
 $r=\nicefrac{R}{64}$, and $s=0$; asynchronous Hyperband loops through brackets $s=0,1,2,3$.  We compare to Vizier without the performance curve early-stopping rule \citep{vizier2017}.\footnote{
At the time of running the experiment, it was  brought to our attention by the team maintaining the Vizier
service that the early-stopping code contained a bug which negatively impacted its performance.  Hence, we omit the results here.  
}   

\begin{figure}[t]
\centering
\begin{subfigure}{0.5\textwidth}
\centering
\includegraphics[width=0.9\textwidth,trim=10 0 10 0]{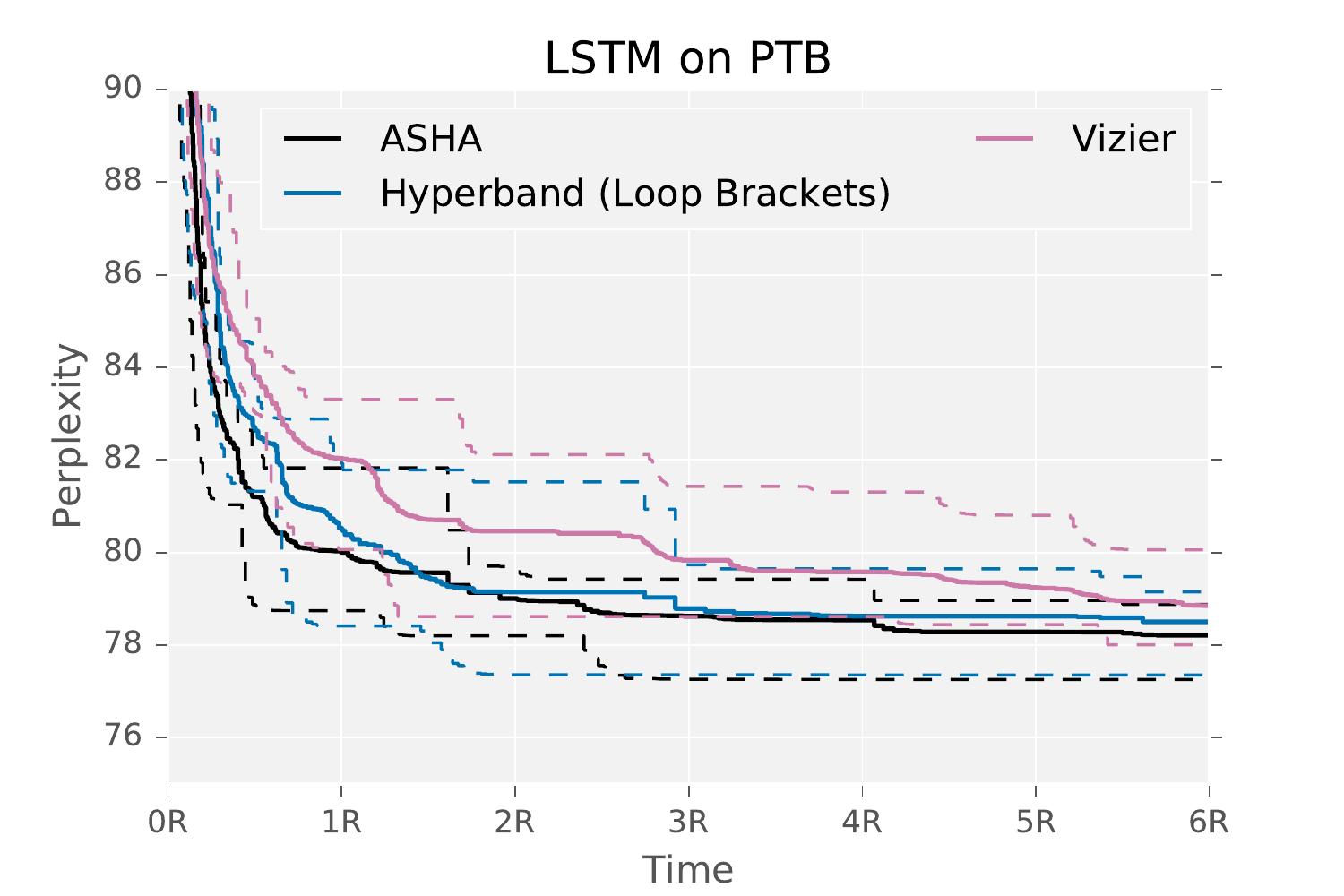}
\end{subfigure}
\caption{\textbf{Large-scale ASHA benchmark} requiring weeks to run with 500 workers.  The x-axis is measured in units of average time to train a single configuration for $R$ resource.  The average across 5 trials is shown, with dashed lines indicating min/max ranges.}
\label{fig:largescale}
\end{figure}

The results in Figure~\ref{fig:largescale} show  that
ASHA and asynchronous Hyperband found good configurations for
this task in $1\times time(R)$.   
Additionally, ASHA and asynchronous Hyperband are both about $3\times$ faster than Vizier at finding a configuration with test perplexity below 80, despite being much simpler and easier to implement.  Furthermore, the best model found by ASHA achieved a test perplexity of 76.6, which is significantly better than 78.4 reported for the large LSTM in \citet{zaremba2014}.
We also note that asynchronous Hyperband initially lags behind ASHA, but eventually catches up at around $1.5\times time(R)$. 

Notably, we observe that certain hyperparameter
configurations in this benchmark induce perplexities that are orders of
magnitude larger than the average case perplexity. Model-based
methods that make assumptions on the data
distribution, such as Vizier, can degrade in performance without further care to
adjust this signal. 
We attempted to alleviate this by capping perplexity scores at 1000 but this still significantly hampered the performance of Vizier.  We view robustness to these types of scenarios as an additional benefit of ASHA
and Hyperband.

\section{Productionizing ASHA}
While integrating ASHA in Determined AI's software platform to deliver production-quality hyperparameter tuning functionality, we encountered several fundamental design decisions that impacted usability, computational performance, and reproducibility.  We next discuss each of these design decisions along with proposed systems optimizations for each decision.

\subsection{Usability}
Ease of use is one of the most important considerations in production; if an advanced method is too cumbersome to use, its benefits may never be realized. In the context of hyperparameter optimization, classical methods like random or grid search require only two intuitive inputs: \emph{number of configurations} ($n$) and \emph{training resources per configuration} ($R$). 
In contrast, as a byproduct of adaptivity, all of the modern methods we considered in this work have many internal hyperparameters. ASHA in particular has the following internal settings: elimination rate $\eta$, early-stopping rate $s$, and, in the case of asynchronous Hyperband, the brackets of ASHA to run.
To facilitate use and increase adoption of ASHA, we simplify its user interface to require the same inputs as random search and grid search, exposing
the internal hyperparameters of ASHA only to advanced users. 

\textbf{Selecting ASHA default settings}. Our experiments in Section~\ref{sec:emp} and the experiments conducted by \citet{LiJamiesonDeSalvoRostamizadehTalwalkar2015} both show
that aggressive early-stopping is effective across a variety of different hyperparameter tuning tasks.  Hence, using both works as guidelines, we propose the following default settings for ASHA: 
\begin{itemize}[leftmargin=*,itemsep=0mm]
\item Elimination rate: we set $\eta = 4$ so that  the top 1/4 of configurations are promoted to the next rung.  
\item Maximum early-stopping rate: we set the maximum early-stopping rate for bracket $s_0$ to allow for a maximum of 5 rungs which indicates a  minimum resource of $r=(\nicefrac{1}{4^4})R=R/256$.  Then the minimum resource per configuration for a given bracket $s$ is $r_s=r\eta^s$.
\item Brackets to run: to increase robustness to misspecification of the early-stopping rate, we default to running the three most aggressively early-stopping brackets $s=0, 1, 2$ of ASHA.  We exclude the two least aggressive brackets (i.e. $s_4$ with $r_4=R$ and $s_3$ with $r_3=R/4$) to allow for higher speedups from early-stopping. 
We define this default set of brackets as the `standard' set of early-stopping brackets, though we also expose the options for more conservative or more aggressive bracket sets.
\end{itemize}

\textbf{Using $n$ as ASHA's stopping criterion}.  Algorithm~\ref{alg:async} does not specify a stopping criterion; instead, it relies on the user to stop once an implicit condition is met, e.g., number of configurations evaluated, compute time, or minimum performance achieved. In a production environment, we decided to use the number of configurations $n$ as an explicit stopping criterion both to match the user interface for random and grid search, and to provide an intuitive connection to the underlying difficulty of the search space.  In contrast, setting a maximum compute time or minimum performance threshold requires prior knowledge that may not be available.

From a technical perspective, $n$ is allocated to the different brackets while maintaining the same total training resources across brackets.  We do this by first calculating the average budget per configuration for a bracket (assuming no incorrect promotions), and then allocating configurations to brackets according to the inverse of this ratio.  For concreteness, let $B$ be the set of brackets we are considering, then the average resource for a given bracket $s$ is $\bar{r}_s=\nicefrac{\text{\# of Rungs}}{\eta^{\lfloor\log_\eta R/r\rfloor-s}}$.
For the default settings described above, this corresponds to $\bar{r}_0=5/256$, $\bar{r}_1=4/64$, and $\bar{r}_2=3/16$, and further translates to $70.5\%$, $22.1\%$, and $7.1\%$ of the configurations being allocated to brackets $s_0, s_1,$ and $s_2$, respectively.  

Note that we still run each bracket asynchronously; the allocated number of configurations $n_s$ for a particular bracket $s$ simply imposes a limit on the width of the bottom rung.
In particular, upon reaching the limit $n_s$ in the bottom rung, the number of pending configurations in the bottom rung is at most equal to the number of workers, $k$.  Therefore, since blocking occurs once a bracket can no longer add configuration to the bottom rung and must wait for promotable configurations, for large-scale problems where $n_s\gg k$, limiting the width of rungs will not block promotions until the bracket is near completion.  In contrast, synchronous SHA is susceptible to blocking from stragglers throughout the entire process, which can greatly reduce both the latency and throughput of configurations promoted to the top rung (e.g. Section~\ref{ssec:smallscale}, Appendix~\ref{appendix:comparison}).
 
\subsection{Automatic Scaling of Parallel Training}
The promotion schedule for ASHA geometrically increases the resource per configuration as we move up the rungs of a bracket.  Hence, the average training time for configurations in higher rungs increases drastically for computation that scales linearly or super-linearly with the training resource, presenting an opportunity to speed up training by using multiple GPUs.  We explore autoscaling of parallel training to exploit this opportunity when resources are available.  

We determine the maximum degree of parallelism for autoscaling a training task using an efficiency criteria motivated by the observation that speedups from parallel training do not scale linearly with the number of GPUs \citep{alexnet2014, inception2014,you2017,lars2017,goyal2017}.  
More specifically, we can use the Paleo framework, introduced by \citet{paleo2017}, to estimate the cost of training neural networks in parallel given different specifications.  \citet{paleo2017} demonstrated that the speedups from parallel training computed using Paleo are fairly accurate when compared to the actual observed speedups for a variety of models.

\begin{figure}[t]
\centering
\begin{subfigure}[h]{6.5cm}
\centering
\includegraphics[width=6.5cm,page=2,trim=10 10 10 10]{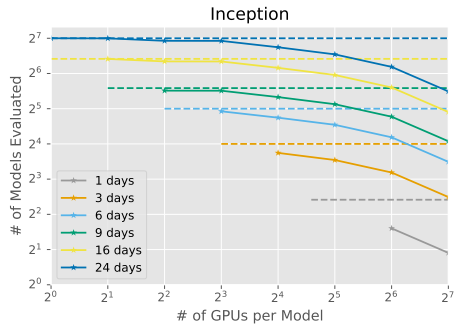}
\end{subfigure}
\caption{\textbf{Tradeoffs for parallel training of Imagenet using Inception V3.}  Given that each configuration takes 24 days to train on a single Tesla K80 GPU, we chart the estimated number (according to the Paleo performance model) of configurations evaluated by 128 Tesla K80s as a function of the number of GPUs used to train each model for different time budgets.  
The dashed line for each color represents the number of 
models evaluated under perfect scaling, i.e. $n$ GPUs train a single model $n$ times as fast, and span the feasible range for number of GPUs per model in order to train within the allocated time budget.  
As expected, more GPUs per configuration are required for smaller time budgets and the total number of configurations evaluated decreases with number of GPUs per model due to decreasing marginal benefit.  
}
\label{fig:modelparallel}
\end{figure}

Figure~\ref{fig:modelparallel} shows Paleo applied to Inception on ImageNet \citep{murray2016} to estimate the training time with different numbers of GPUs
under strong scaling (i.e. fixed batch size with increasing
parallelism), Butterfly AllReduce communication  scheme, specified
hardware settings (namely Tesla K80 GPU and 20G Ethernet), and a batch
size of 1024. 

The diminishing returns when using more GPUs to train a single model is evident in Figure~\ref{fig:modelparallel}.  Additionally, there is a tradeoff between
using resources to train a model faster to reduce latency versus evaluating more configurations to increase throughput. Using the predicted tradeoff curves generated using Paleo, we can automatically limit the number of GPUs per configuration to control efficiency relative to perfect linear scaling, e.g., if the desired level of efficiency is at least $75\%$, then we would limit the number of GPUs per configuration for Inception to at most 16 GPUs. 

\subsection{Resource Allocation}
\label{ssec:resource_all}
Whereas research clusters often require 
users to specify the number of workers requested and allocate workers on a first-in-first-out (FIFO) fashion, this scheduling mechanism is poorly suited for production settings for two main reasons.  First, as we discuss below in the context of ASHA, machine learning workflows can have variable resource requirements over the lifetime of a job, and forcing users to specify static resource requirements can result in suboptimal cluster utilization.  Second, 
FIFO scheduling can result in poor sharing of cluster resources among users, as a single large job could saturate the cluster and block all other user jobs.

We address these issues with a centralized fair-share scheduler that adaptively allocates resources over the lifetime of each job. 
Such a scheduler must both (i) determine the appropriate amount of parallelism for each individual job, and (ii) allocate computational resources across all user jobs. 
In the context of an ASHA workload, the scheduler automatically determines the maximum resource requirement at any given time based on the inputs to ASHA and the parallel scaling profile determined by Paleo.  Then, the scheduler allocates cluster resources 
by considering the resource requirements of all jobs while maintaining fair allocation across users.  We describe each of these components in more detail below.

\textbf{Algorithm level resource allocation.}
Recall that we propose to use the number of configurations, $n$, as a stopping criteria for ASHA in production settings.  Crucially, this design decision limits the maximum degree of parallelism for an ASHA job.
If $n$ is the number of desired configurations for a given ASHA bracket  and $\kappa$ the maximum allowable training parallelism, e.g., as determined by Paleo, then at initialization, the maximum parallelism for the bracket is $n\kappa$.  We maintain a stack of training tasks $\mathbf{S}$ that is populated initially with all configurations for the bottom rung $n$.  The top task in $\mathbf{S}$ is popped off whenever a worker requests a task and promotable configurations are added to the top of $\mathbf{S}$ when tasks complete.  As ASHA progresses, the maximum parallelism is adaptively defined as $\kappa|\mathbf{S}|$.  Hence, an adaptive worker allocation schedule that relies on $\kappa|\mathbf{S}|$ would improve cluster utilization relative to a static allocation scheme, without adversely impacting performance.  

\textbf{Cluster level resource allocation.}
Given the maximum degree of parallelism for any ASHA job, the scheduler then allocates resources uniformly across all jobs 
while respecting these maximum parallelism limits.  
We allow for an optional priority weighting factor so that certain jobs can receive a larger ratio of the total computational resources.  Resource allocation is performed using a water-filling scheme where any allocation above the maximum resource requirements for a job are distributed evenly to remaining jobs.  

For concreteness, consider a scenario in which we have a cluster of 32 GPUs shared between a group of users.  When a single user is running an ASHA job with 8 configurations in $\mathbf{S_1}$ and a maximum training parallelism of $\kappa_1=4$, the scheduler will allocate all 32 GPUs to this ASHA job.  When another user submits an ASHA job with a maximum parallelism of $\kappa_2|\mathbf{S_2}|=64$, the central scheduler will then allocate 16 GPUs to each user.  This simple scenario demonstrate how our central scheduler allows jobs to benefit from maximum parallelism when the computing resources are available, while maintaining fair allocation across jobs in the presence of resource contention.  

\subsection{Reproducibility in Distributed Environments}
Reproducibility is critical in production settings to instill trust during the model development process;  foster collaboration and knowledge transfer across teams of users; and  allow for fault tolerance and iterative refinement of models.  However, ASHA introduces two primary reproducibility challenges, each of which we describe below.

\textbf{Pausing and restarting configurations}.  There are many sources of randomness when training machine learning models; some source can be made deterministic by setting the random seed, while others related to GPU floating-point computations and CPU multi-threading are harder to avoid without performance ramifications.  Hence, reproducibility when resuming promoted configurations requires carefully checkpointing all stateful objects pertaining to the model.  At a minimum this includes the model weights, model optimizer state, random number generator states, and data generator state.  We provide a checkpointing solution that facilitates reproducibility in the presence of stateful variables and seeded random generators.  The availability of deterministic GPU floating-point computations is dependent on the deep learning framework, but we allow users to control for all other sources of randomness during training.  

\textbf{Asynchronous promotions}. 
To allow for full reproducibility of ASHA, we track the sequence of all promotions made within a bracket.  This sequence fixes the nondeterminism from asynchrony, allowing subsequent replay of the exact promotions as the original run.
Consequently, we can reconstruct the full state of a bracket at any point in time, i.e. which configurations are on which rungs and which training tasks are in the stack.  

Taken together, reproducible checkpoints and full bracket states allow us to seamlessly resume hyperparameter tuning jobs when crashes happen and allow users to request to evaluate more configurations if desired.  
For ASHA, refining hyperparameter selection by resuming an existing bracket is highly beneficial, since a wider rung gives better empirical estimates of the top $1/\eta$ configurations.

\section{Conclusion}
In this paper, we addressed the problem of developing a production-quality system for hyperparameter tuning by introducing ASHA, a theoretically principled method for simple and robust massively parallel hyperparameter optimization.  We presented empirical results demonstrating that ASHA outperforms state-of-the-art methods Fabolas, PBT, BOHB, and Vizier in a suite of hyperparameter tuning benchmarks.   Finally, we provided systems level solutions to improve the effectiveness of ASHA that are applicable to existing systems that support our algorithm.   

\section*{Acknowledgements}
This work was supported in part by DARPA FA875017C0141, the National Science Foundation grants IIS1705121 and IIS1838017, an Okawa Grant, a Google Faculty Award, an Amazon Web Services Award, a JP Morgan A.I. Research Faculty Award, and a Carnegie Bosch Institute Research Award. Any opinions, findings and conclusions or recommendations expressed in this material are those of the author(s) and do not necessarily reflect the views of DARPA, the National Science Foundation, or any other funding agency.
\bibliography{hyperband}
\bibliographystyle{mlsys2020}

\clearpage
\appendix
\section{Appendix}
As part of our supplementary material, (1) compare the impact of stragglers and dropped jobs on synchronous SHA and ASHA, (2) present the comparison to Fabolas in the sequential setting and (3) provide additional details for the empirical results shown in Section~\ref{sec:emp}.

\subsection{Comparison of Synchronous SHA and ASHA}
\label{appendix:comparison}
\begin{figure*}[t]
\begin{subfigure}{0.48\textwidth}
\centering
\includegraphics[width=\textwidth,page=1]{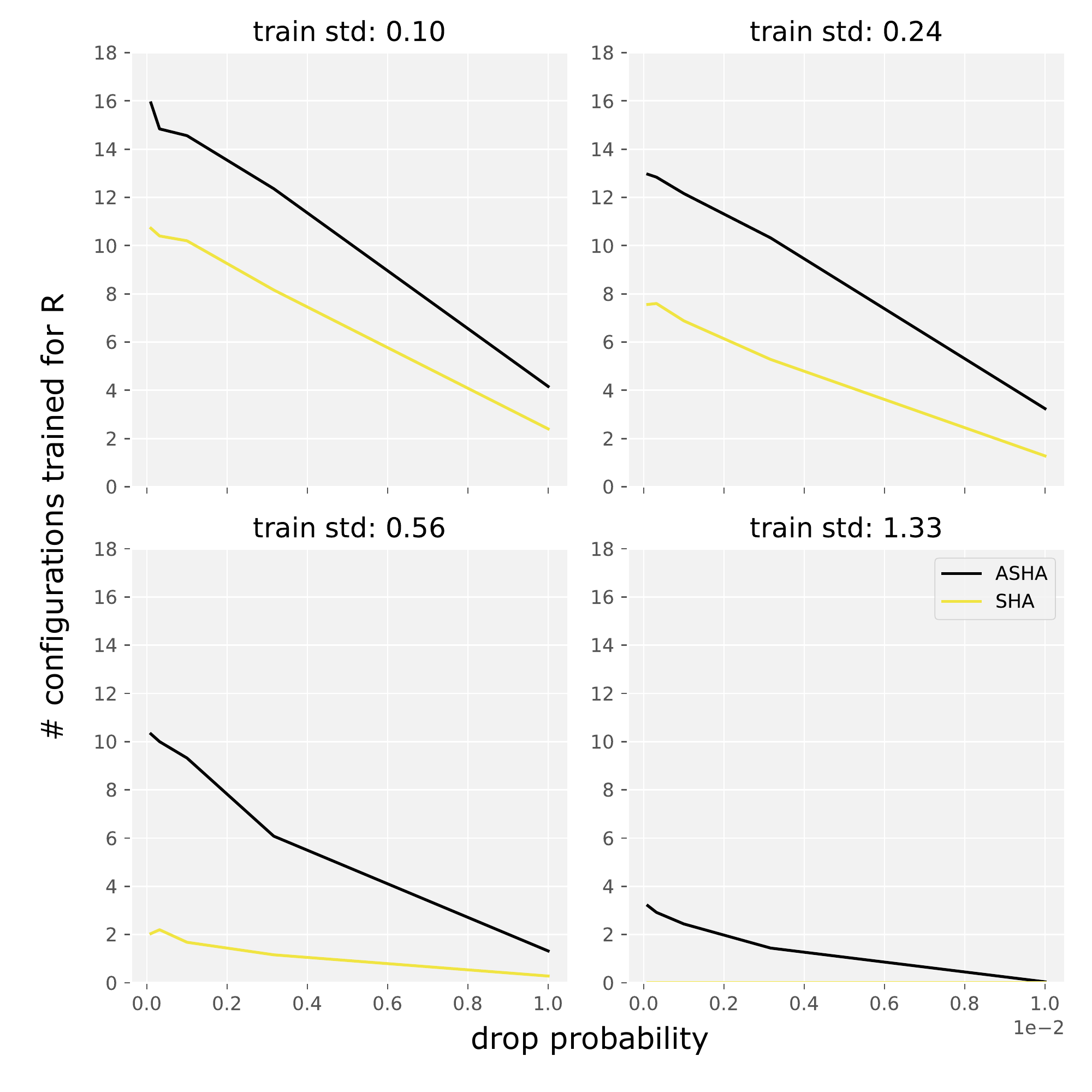}
\caption{Average number of configurations trained on $R$ resource.}
\label{fig:simncompleted}
\end{subfigure}
\hspace{0.5cm}
\begin{subfigure}{0.48\textwidth}
\centering
\includegraphics[width=\textwidth,page=1]{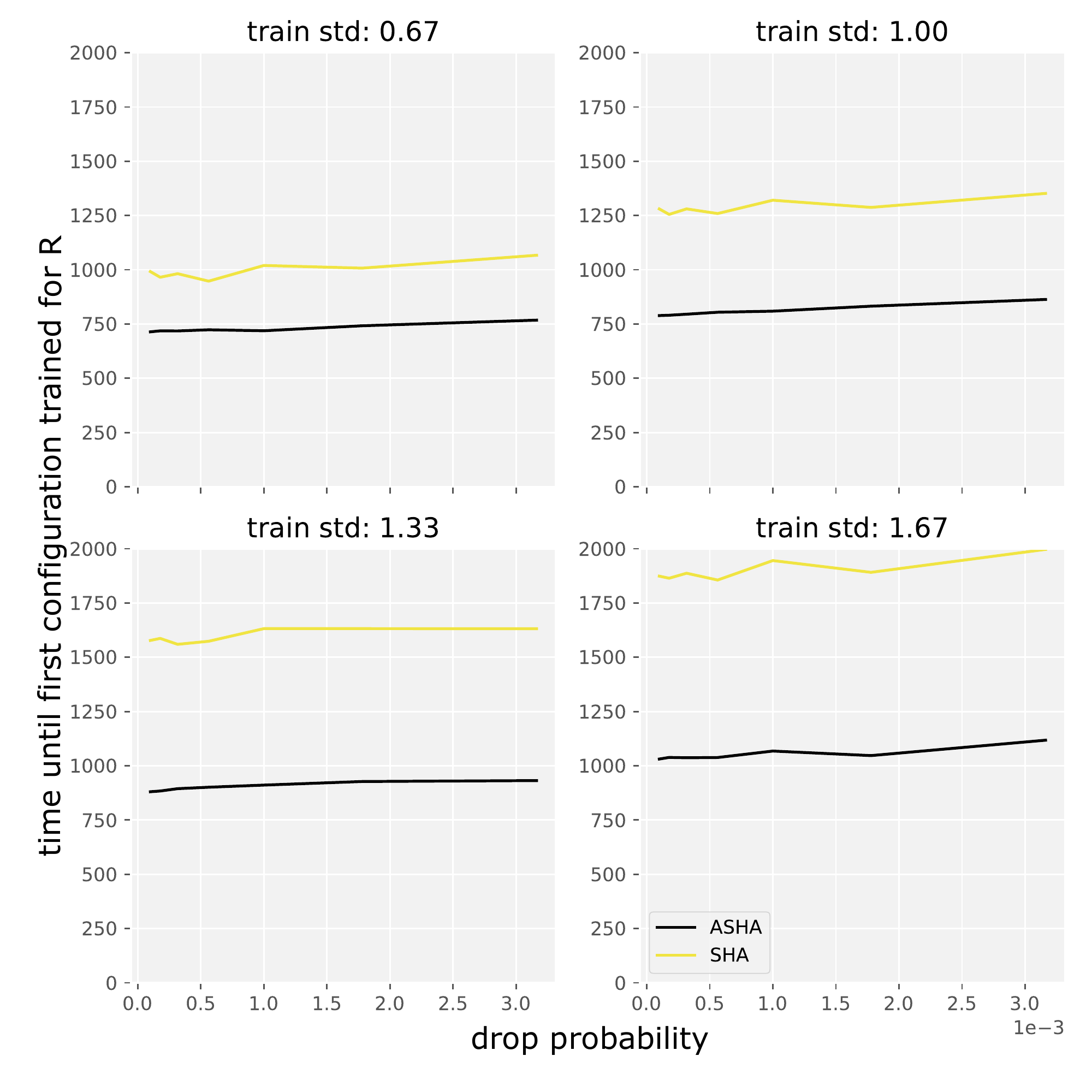}
\caption{Average time before a configuration is trained on $R$ resource.}
\label{fig:simtcompleted}
\end{subfigure}
\caption{\textbf{Simulated workloads comparing impact of stragglers and dropped jobs.} The number of configurations trained for $R$ resource (left) is higher for ASHA than synchronous SHA when the standard deviation is high.  Additionally, the average time before a configuration is trained for $R$ resource (right) is lower for ASHA than for synchronous SHA when there is high variability in training time (i.e., stragglers). Hence, ASHA is more robust to stragglers and dropped jobs than synchronous SHA since it returns a completed configuration faster and returns more configurations trained to completion.}
\label{fig:sim}
\end{figure*}
We use simulated workloads to evaluate the impact of stragglers and dropped jobs on synchronous SHA and ASHA.  For our simulated workloads, we run synchronous SHA with $\eta=4$, $r=1$, $R=256$, and $n=256$ and ASHA with the same values and the maximum early-stopping rate $s=0$.  Note that BOHB \citep{falkner2018bohb}, one of the competitors we empirically compare to in Section~\ref{sec:emp}, is also susceptible to stragglers and dropped jobs since it uses synchronous SHA as its parallelization scheme but leverages Bayesian optimization to perform adaptive sampling.  

For these synthetic experiments, we assume that the expected training time for each job is the same as the allocated resource.  We simulate stragglers by multiplying the expected training time by $(1+|z|)$ where $z$ is drawn from a normal distribution with mean 0 and a specified standard deviation.  We simulated dropped jobs by assuming that there is a given $p$ probability that a job will be dropped at each time unit, hence, for a job with a runtime of 256 units, the probability that it is not dropped is $1- (1-p)^{256}$.    

Figure~\ref{fig:sim} shows the number of configurations trained to completion (left) and time required before one configuration is trained to completion (right) when running synchronous SHA and ASHA using 25 workers. For each combination of training time standard deviation and drop probability, we simulate ASHA and synchronous SHA 25 times and report the average. As can be seen in Figure~\ref{fig:simncompleted}, ASHA trains many more configurations to completion than synchronous SHA when the standard deviation is high; we hypothesize that this is one reason ASHA performs significantly better than synchronous SHA and BOHB for the second benchmark in Section~\ref{ssec:smallscale}.  Figure~\ref{fig:simtcompleted} shows that ASHA returns a configuration trained for the maximum resource $R$ much faster than synchronous SHA when there is high variability in training time (i.e., stragglers) and high risk of dropped jobs.  Although ASHA is more robust than synchronous SHA to stragglers and dropped jobs on these simulated workloads, we nonetheless compare synchronous SHA in Section~\ref{ssec:largescale} and show that ASHA performs better.  

\subsection{Comparison with Fabolas in Sequential Setting}	
\label{ssec:fabolas}
\begin{figure*}[t]
\centering
\begin{subfigure}{0.48\textwidth}
\centering
\includegraphics[width=\textwidth,page=1,trim=10 0 10 0]{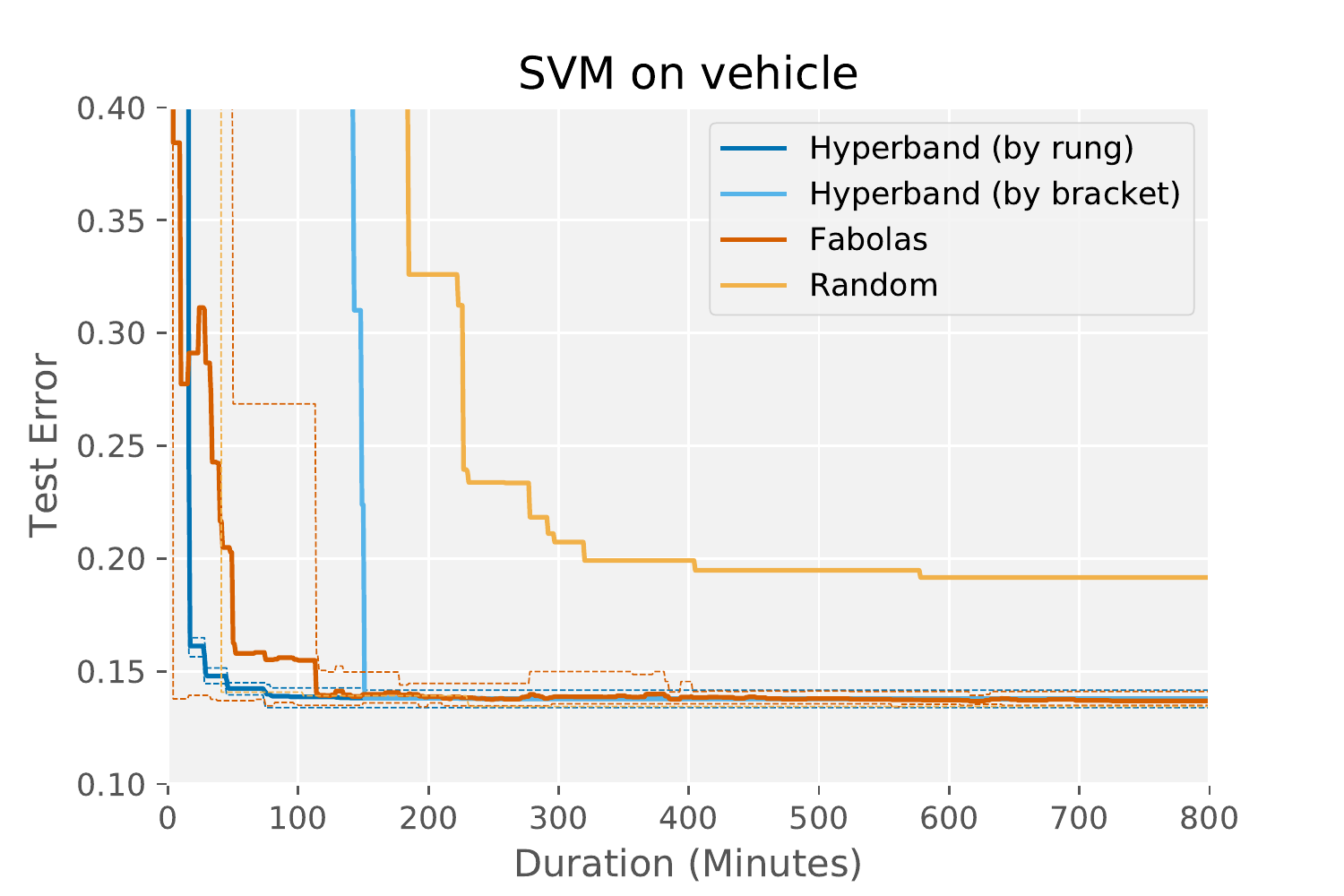}
\end{subfigure}
\begin{subfigure}{0.48\textwidth}
\centering
\includegraphics[width=\textwidth,page=2,trim=10 0 10 0]{newcharts.pdf}
\end{subfigure}
\begin{subfigure}{0.48\textwidth}
\centering
\includegraphics[width=\textwidth,page=3,trim=10 0 10 0]{newcharts.pdf}
\end{subfigure}
\begin{subfigure}{0.48\textwidth}
\centering
\includegraphics[width=\textwidth,page=4,trim=10 0 10 0]{newcharts.pdf}
\end{subfigure}

\caption{\textbf{Sequential Experiments} (1 worker) with Hyperband running synchronous SHA.  Hyperband (by rung) records the incumbent after the completion of a SHA rung, while Hyperband (by bracket) records the incumbent after the completion of an entire SHA bracket.  The average test error across 10 trials of each hyperparameter optimization method is shown in each plot.  Dashed lines represent min and max ranges for each tuning method.   } \label{fig:fabolas}
\end{figure*}

\cite{fabolas2016} showed that Fabolas can be over an order of magnitude faster than existing Bayesian optimization
 methods.  
Additionally, the empirical studies presented in \cite{fabolas2016} suggest that Fabolas is faster than Hyperband at finding a good configuration. 
 We conducted our own experiments to compare Fabolas with Hyperband on the following tasks: 
 \begin{enumerate}
 \item Tuning an SVM using the same search space as \cite{fabolas2016}.
 \item Tuning a convolutional neural network (CNN) with the same search space as \citet{hyperband} on CIFAR-10 \citep{cifar10data}.
 \item Tuning a CNN on SVHN \citep{svhndata} with varying number of layers, batch size, and number of filters (see Appendix~\ref{smallcnn} for more details).
 \end{enumerate}
In the case of the SVM task, the allocated resource is number of training datapoints, while for the CNN tasks, the allocated
resource is the number of training iterations. 

We note that Fabolas was specifically designed for data points as the resource, and hence, is not directly applicable to tasks (2) and (3).  However, freeze-thaw Bayesian optimization \citep{swersky2014freeze}, which was specifically designed for models that use iterations as the resource, is known to perform poorly on deep learning tasks \citep{earlystopping2015}.  Hence, we believe Fabolas to be a reasonable competitor for tasks (2) and (3) as well, despite the aforementioned shortcoming. 

We use the same evaluation framework as \citet{fabolas2016}, where the best configuration, also known as the {\em incumbent}, is recorded through time and the test error is calculated in an offline validation step.  Following \citet{fabolas2016}, the incumbent for Hyperband is taken to be the configuration with the lowest validation loss and the incumbent for Fabolas is the configuration with the lowest predicted validation loss on the full dataset.   Moreover, for these experiments, we set $\eta=4$ for Hyperband.

Notably, when tracking the best performing configuration for Hyperband, we consider two approaches.  We first consider the approach proposed in \citet{LiJamiesonDeSalvoRostamizadehTalwalkar2015} and used by~\citet{fabolas2016} in their evaluation of Hyperband.  In this variant, which we refer to as ``Hyperband (by bracket),'' the incumbent is 
recorded \emph{after the completion of each SHA bracket}. 
We also consider a second approach where we
record the incumbent \emph{after the completion of each rung} of SHA to make use of intermediate validation losses, similar to what we propose for ASHA (see discussion in Section~\ref{algo_discussion} for details).  We will refer to 
Hyperband using this accounting scheme as ``Hyperband (by rung).''  Interestingly, by leveraging these intermediate losses, we observe that Hyperband actually outperforms Fabolas.   

In Figure~\ref{fig:fabolas}, we show the performance of Hyperband, Fabolas, and random search.  
Our results show that Hyperband (by rung) is competitive with Fabolas at finding a good configuration and will often find a better configuration than Fabolas with less variance.  Note that Hyperband loops through the brackets of SHA, ordered by decreasing early-stopping rate; the first bracket finishes when the test error for Hyperband (by bracket) drops.  Hence, most of the progress made by Hyperband comes from the bracket with the most aggressive early-stopping rate, i.e. bracket 0. 

\subsection{Experiments in Section~\ref{ssec:sequential} and Section~\ref{ssec:smallscale}}
\label{appendix:pbt}
We use the usual train/validation/test splits for CIFAR-10, evaluate configurations on the validation set to inform algorithm decisions, and report test error.  These experiments were conducted using \texttt{g2.2xlarge} instances on Amazon AWS. 

For both benchmark tasks, we run SHA and BOHB with $n=256$, $\eta=4$, $s=0$, and set $r=R/256$, where $R=30000$ iterations of stochastic gradient descent.  Hyperband loops through 5 brackets of SHA, moving from bracket $s=0, r=R/256$ to bracket $s=4, r=R$.  We run ASHA and asynchronous Hyperband with the same settings as the synchronous versions.   We run PBT with a population size of 25, which is between the recommended 20--40 \citep{jaderberg2017pbt}.  Furthermore, to help PBT evolve from a good set of configurations, we randomly sample configurations until at least half of the population performs above random guessing. 

 We implement PBT with truncation selection for the exploit phase, where the bottom $20\%$ of configurations are replaced with a uniformly sampled configuration from the top $20\%$ (both weights and hyperparameters are copied over).  Then, the inherited hyperparameters pass through an exploration phase where $\nicefrac{3}{4}$ of the time they are either perturbed by a factor of 1.2 or 0.8 (discrete hyperparameters are perturbed to two adjacent choices), and $\nicefrac{1}{4}$ of the time they are randomly resampled.  Configurations are considered for exploitation/exploration every 1000 iterations, for a total of 30 rounds of adaptation.  For the experiments in Section~\ref{ssec:smallscale}, to maintain 100\% worker efficiently for PBT while enforcing that all configurations are trained for within 2000 iterations of each other, we spawn new populations of 25 whenever a job is not available from existing populations.
\begin{table}[h]
\centering
\begin{tabular}{c c c} 
 Hyperparameter & Type & Values \\
 \hline 
 \\ [-1em]
 batch size & choice & $\{2^6, 2^7, 2^8, 2^9\}$ \\
 \# of layers & choice & $\{2, 3, 4\}$ \\
 \# of filters & choice & $\{16, 32, 48, 64\}$ \\
 weight init std 1 & continuous & $\log \; [10^{-4},10^{-1}]$ \\
 weight init std 2 & continuous & $\log \; [10^{-3},1]$ \\
 weight init std 3 & continuous & $\log \; [10^{-3},1]$ \\
$l_2$ penalty 1 &  continuous & $\log \; [10^{-5},1]$ \\
$l_2$ penalty 2 & continuous & $\log \; [10^{-5},1]$ \\
$l_2$ penalty 3 & continuous & $\log \; [10^{-3},10^{2}]$ \\
 learning rate & continuous & $\log \; [10^{-5},10^{1}]$ \\
\end{tabular}
\caption{Hyperparameters for small CNN architecture tuning task.}
\label{table:smallcnn}
\end{table}
 
Vanilla PBT is not compatible with hyperparameters that change the architecture of the neural network, since inherited weights are no longer valid once those hyperparameters are perturbed.  To adapt PBT for the architecture tuning task, we fix hyperparameters that affect the architecture in the
explore stage.  Additionally, we restrict configurations to be trained within 2000 iterations of each other so a fair comparison is made to select configurations to exploit.  If we do not impose this restriction, PBT will be biased against configurations that take longer to train, since it will be comparing these configurations with those that have been trained for more iterations.  

\subsection{Experimental Setup for the Small CNN Architecture Tuning Task}
\label{smallcnn}

This benchmark tunes a multiple layer CNN network with the hyperparameters shown in Table~\ref{table:smallcnn}.  This search space was used for the small architecture task on SVHN (Section~\ref{ssec:fabolas}) and CIFAR-10 (Section~\ref{ssec:smallscale}).  The \# of layers hyperparameter indicate the number of convolutional layers before two fully connected layers.  The \# of filters indicates the \# of filters in the CNN layers with the last CNN layer having $2\times \text{\# filters}$.  Weights are initialized randomly from a Gaussian distribution with the indicated standard deviation.  There are three sets of weight init and $l_2$ penalty hyperparameters; weight init 1 and $l_2$ penalty 1 apply to the convolutional layers, weight init 2 and $l_2$ penalty 2 to the first fully connected layer, and weight init 3 and $l_2$ penalty 3 to the last fully connected layer.  Finally, the learning rate hyperparameter controls the initial learning rate for SGD.  All models use a fixed learning rate schedule with the learning rate decreasing by a factor of 10 twice in equally spaced intervals over the training window.  This benchmark is run on the SVHN dataset \citep{svhndata} following \citet{svhnsplit} to create the train, validation, and test splits.  

\subsection{Experimental Setup for Neural Architecture Search Benchmarks}
\label{app:nas}
For NAS benchmarks evaluated in Section~\ref{ssec:nas}, we used the same search space as that considered by \citet{liu2019darts} for designing CNN and RNN cells.  Following \citet{litalwalkar2019}, we sample architectures from the associated search space randomly and train them using the same hyperparameter settings as that used by \citet{liu2019darts} in the evaluation stage.  We refer the reader to the following code repository for more details:
\url{https://github.com/liamcli/darts_asha}.

\subsection{Tuning Modern LSTM Architectures}
\label{ssec:sota_lstm}

\begin{figure}[t]
\centering
\begin{subfigure}{0.5\textwidth}
\centering
\includegraphics[width=0.9\textwidth,trim=10 0 10 0]{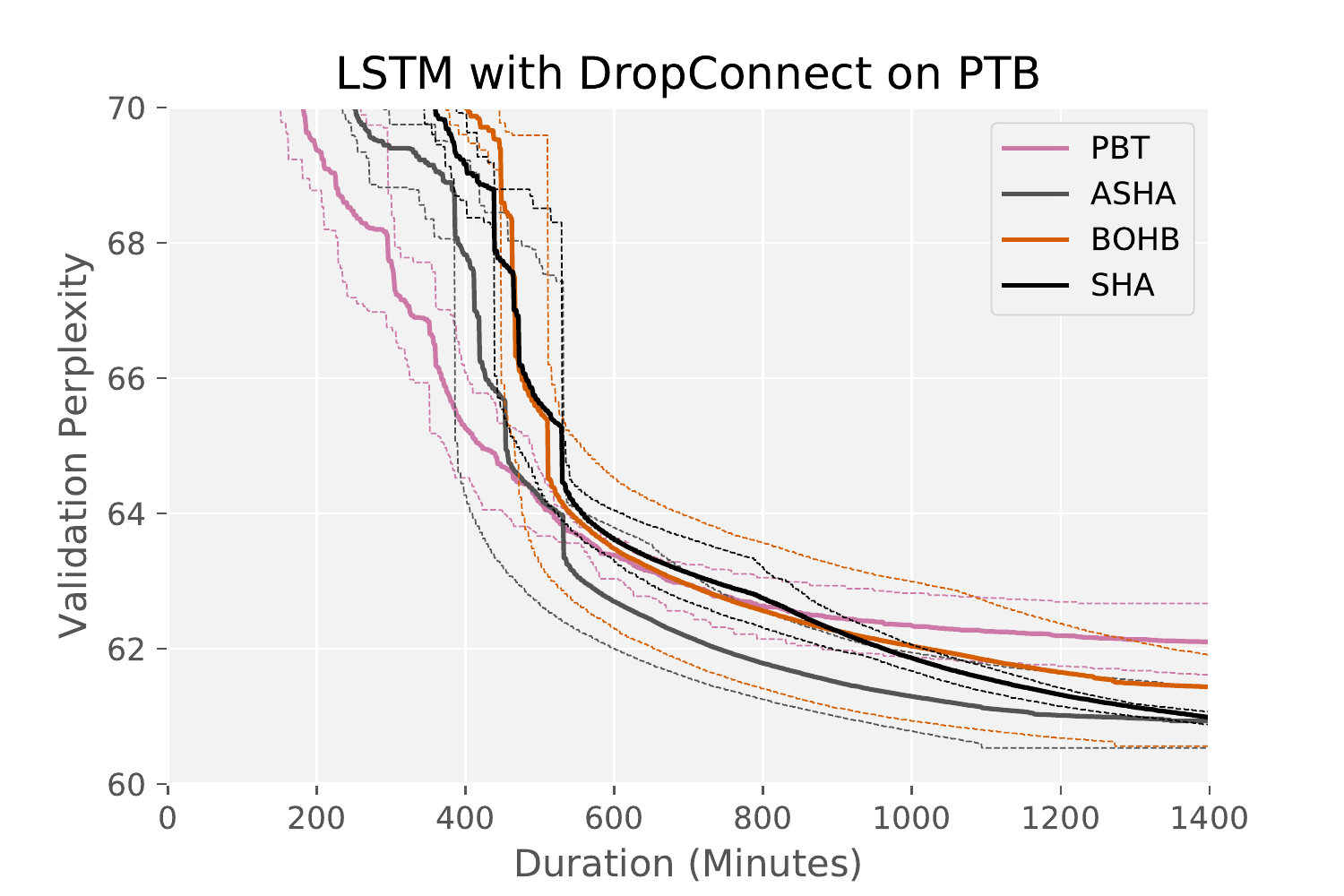}
\end{subfigure}
\caption{\textbf{Modern LSTM benchmark} with DropConnect \citep{merity2018awd} using 16 GPUs.  The average across 5 trials is shown, with dashed lines indicating min/max ranges.}
\label{fig:sota_ptb}
\end{figure}

As a followup to the experiment in Section~\ref{ssec:nas},
we consider a search space for language modeling that is able to achieve near state-of-the-art performance.  Our starting point was the work of \citet{merity2018awd}, which introduced a near state-of-the-art LSTM architecture with a more effective regularization scheme called DropConenct.  We constructed a search space around their configuration and ran ASHA and PBT, each with 16 GPUS on one \texttt{p2.16xlarge} instance on AWS. 
The hyperparameters that we considered along with their associated ranges are shown in Table~\ref{tab:sota_lstm}.

Then, for ASHA, SHA, and BOHB we used $\eta=4$, $r=1$ epoch,  $R=256$ epochs, and $s=0$.  For PBT, we use a population size to 20, a maximum resource of $256$ epochs, and perform explore/exploit every 8 epochs using the same settings as the previous experiments.  

Figure~\ref{fig:sota_ptb} shows that while PBT performs better initially, ASHA soon catches up and finds a better final configuration; in fact, the min/max ranges for ASHA and PBT do not overlap at the end.  We then trained the best configuration found by ASHA for more epochs and reached validation and test perplexities of 60.2 and 58.1 respectively before fine-tuning and 58.7 and 56.3 after fine-tuning.  For reference, \citet{merity2018awd} reported validation and test perplexities respectively of 60.7 and 58.8 without fine-tuning and 60.0 and 57.3 with fine-tuning. This demonstrates the effectiveness of ASHA in the large-scale regime for modern hyperparameter optimization problems.

\begin{table}[h]
\centering
\begin{tabular}{c c c} 
 Hyperparameter & Type & Values \\ 
 \hline
learning rate & continuous & $\log\;[10,100]$ \\
dropout (rnn) & continuous & $[0.15, 0.35]$\\
dropout (input) & continuous & $[0.3,0.5]$ \\
dropout (embedding) & continuous & $[0.05, 0.2]$ \\
dropout (output) & continuous & $[0.3, 0.5]$\\
dropout (dropconnect) & continuous & $[0.4, 0.6]$ \\
weight decay & continuous & $\log \; [0.5e-6,2e-6]$\\
batch size & discrete & $[15, 20, 25]$\\
time steps & discrete & $[65, 70, 75]$
\end{tabular}
\caption{Hyperparameters for 16 GPU near state-of-the-art LSTM task.}
\label{tab:sota_lstm}
\end{table}

\subsection{Experimental Setup for Large-Scale Benchmarks}
\label{appendix:largescale}

\begin{table}[h]
\centering
\begin{tabular}{c c c} 
 Hyperparameter & Type & Values \\ 
 \hline
 batch size & discrete & $[10, 80]$ \\
 \# of time steps & discrete & $[10, 80]$ \\
 \# of hidden nodes & discrete & $[200, 1500]$ \\
 learning rate & continuous & $\log \; [0.01, 100.]$ \\
 decay rate & continuous & $[0.01, 0.99]$\\
decay epochs & discrete & $[1, 10]$ \\
 clip gradients & continuous & $[1, 10]$ \\
 dropout probability & continuous & $[0.1, 1.]$ \\
 weight init range & continuous & $\log \; [0.001, 1]$
\end{tabular}
\caption{Hyperparameters for PTB LSTM task.}
\label{table:ptb}
\end{table}

The hyperparameters for the LSTM tuning task comparing ASHA to Vizier on the Penn Tree Bank (PTB) dataset presented in Section~\ref{ssec:largescale} is shown in Table~\ref{table:ptb}.  Note that all
hyperparameters are tuned on a linear scale and sampled uniform over the specified range.  The inputs to the LSTM layer are embeddings of the words in a sequence.  The number of hidden nodes hyperparameter refers to the 
number of nodes in the LSTM.  The learning rate is decayed by the decay rate after each interval of decay steps.  Finally, the weight initialization range indicates the upper bound of the uniform distribution used to initialize all weights.  The other hyperparameters
have their standard interpretations for neural networks.  The default training (929k words) and test (82k words) splits for PTB are used for training and evaluation \citep{ptb}.   We define resources as the number of training records, which translates into
the number of training iterations after accounting for certain hyperparameters.

\end{document}